\documentclass[conference]{IEEEtran}
\IEEEoverridecommandlockouts
\usepackage{cite}
\usepackage{amsmath,amssymb,amsfonts}
\usepackage{algorithmic}
\usepackage{graphicx}
\usepackage{textcomp}
\usepackage{xcolor}
\usepackage{booktabs}
\usepackage{soul}
\def\BibTeX{{\rm B\kern-.05em{\sc i\kern-.025em b}\kern-.08em
    T\kern-.1667em\lower.7ex\hbox{E}\kern-.125emX}}
\begin{document}

\title{Average-Over-Time Spiking Neural Networks for Uncertainty Estimation in Regression\\
\thanks{Tao Sun is supported by NWO-NWA grant NWA.1292.19.298.}
}


\author{\IEEEauthorblockN{Tao Sun\IEEEauthorrefmark{1}, Sander Boht\'e \IEEEauthorrefmark{2}}

\IEEEauthorblockA{\IEEEauthorrefmark{1}\IEEEauthorrefmark{2}Machine Learning Group,
CWI, Amsterdam, The Netherlands\\
\IEEEauthorrefmark{2} Cognitive and Systems Neuroscience Group, University of Amsterdam, Amsterdam, The Netherlands \\
Email: \IEEEauthorrefmark{1}tao.sun@cwi.nl, \IEEEauthorrefmark{3}sbohte@cwi.nl}}

\maketitle

\begin{abstract}
Uncertainty estimation is a standard tool to quantify the reliability of modern deep learning models, and crucial for many real-world applications. However, efficient uncertainty estimation methods for spiking neural networks, particularly for regression models, have been lacking. Here, we introduce two methods that adapt the Average-Over-Time Spiking Neural Network (AOT-SNN) framework to regression tasks, enhancing uncertainty estimation in event-driven models. The first method uses the heteroscedastic Gaussian approach, where SNNs predict both the mean and variance at each time step, thereby generating a conditional probability distribution of the target variable. The second method leverages the Regression-as-Classification (RAC) approach, reformulating regression as a classification problem to facilitate uncertainty estimation. We evaluate our approaches on both a toy dataset and several benchmark datasets, demonstrating that the proposed AOT-SNN models achieve performance comparable to or better than state-of-the-art deep neural network methods, particularly in uncertainty estimation. Our findings highlight the potential of SNNs for uncertainty estimation in regression tasks, providing an efficient and biologically inspired alternative for applications requiring both accuracy and energy efficiency.
\end{abstract}


\section{Introduction}
Uncertainty estimation plays a pivotal role in machine learning, especially in high-stakes applications such as autonomous systems, healthcare, and financial forecasting, where the confidence in predictions is as vital as the predictions themselves \cite{yarin2016uncertainty}. In these high-stake models, uncertainty is often represented through probabilities, with high-quality uncertainty estimation ensuring that predicted probabilities accurately reflect the true likelihood of a prediction being correct \cite{ghahramani2015probabilistic}. While classification models typically provide a probability distribution over discrete classes, which facilitates uncertainty estimation, regression models produce continuous outputs, making uncertainty estimation significantly more challenging \cite{lakshminarayanan2017simple}. In deep neural networks (DNNs), uncertainty estimation has been extensively studied for both classification and regression tasks, thereby enhancing model reliability and robustness \cite{gawlikowski2023survey,abdar2021review}. However, the methods developed for uncertainty estimation in DNNs are generally not directly applicable to event-driven neural networks, such as spiking neural networks (SNNs) \cite{sun2023efficient}.

\begin{figure*}[!t]
  \centering
     \includegraphics[width=0.85\textwidth]{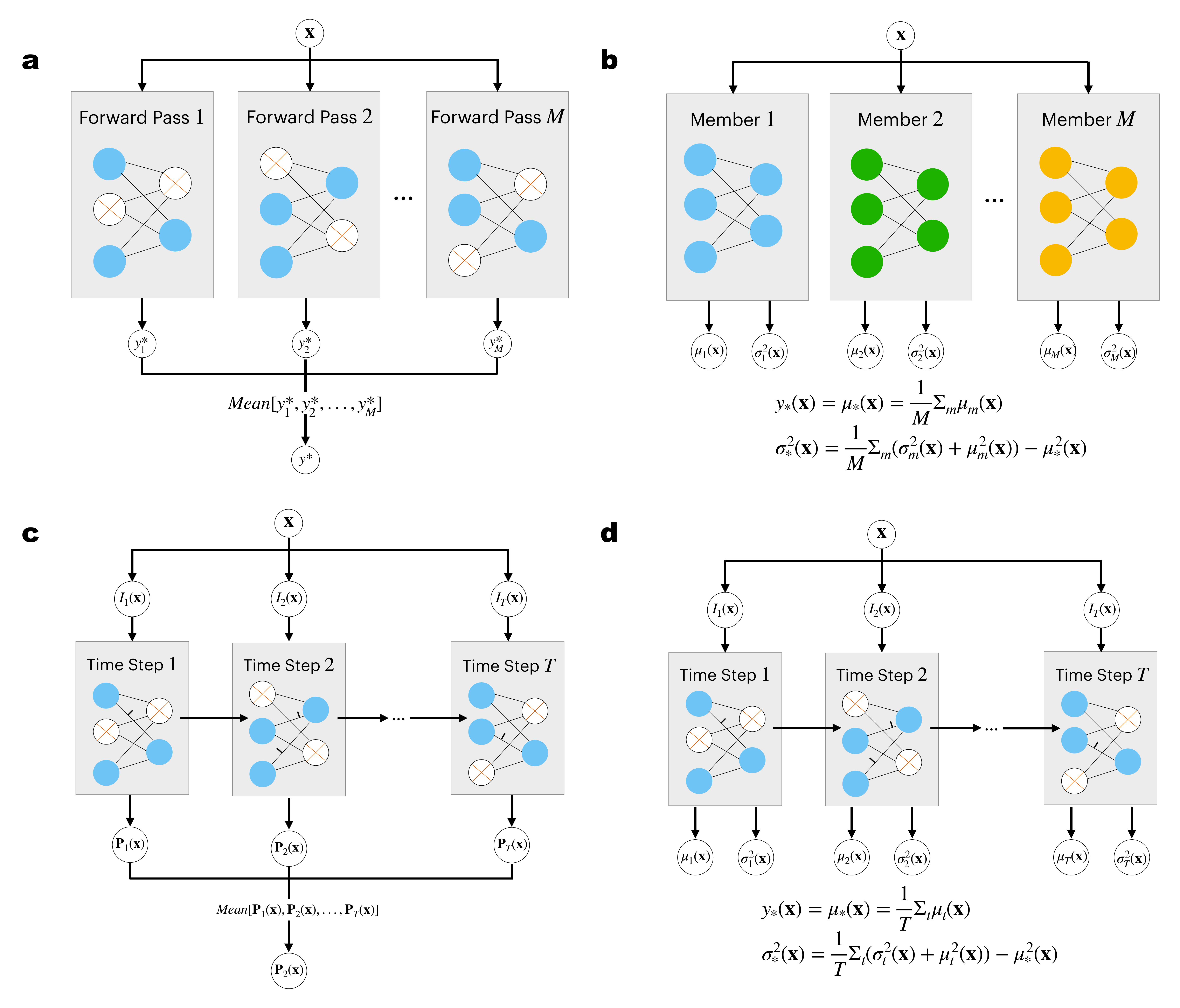}
  \caption{\textbf{a}, The MC-dropout method estimates uncertainty by averaging the outputs of multiple forward passes through a dropout-enabled DNN. \textbf{b}, Deep ensembles train an ensemble of DNNs with identical architectures. For regression tasks, each member in an ensemble assumes a heteroscedastic Gaussian distribution and outputs two variables per input: the mean $\mu_m(\mathbf{x})$ and variance $\sigma_m^2(\mathbf{x})$. The predictive mean $\mu_*(\mathbf{x})$ and variance $\sigma_*(\mathbf{x})^2$ of the ensemble are derived from these two variables. \textbf{c}, In classification, AOT-SNNs apply continual MC-dropout in SNNs by averaging the probabilities across time steps to generate the predictive distribution. \textbf{d}, In regression, at each time step, an AOT-SNN outputs the the mean $\mu_t(\mathbf{x})$ and variance $\sigma_t^2(\mathbf{x})$, with the final predictive mean $\mu_*(\mathbf{x})$ and variance $\sigma_*(\mathbf{x})^2$ calculated from these values.}
  \label{fig:arch_detail}
\end{figure*}

Inspire by biological computation, SNNs are increasingly gaining attention for real-time, resource-constrained applications \cite{schuman2022opportunities}. The sparse and asynchronous event-driven nature of SNNs makes them highly suitable for tasks requiring low-latency decision-making and energy-efficient processing \cite{sun2024dpsnn}. However, these characteristics also pose considerable challenges for integrating uncertainty estimation techniques originally developed for DNNs. For DNNs, Monte Carlo dropout (MC-dropout) \cite{gal2016dropout} and deep ensembles \cite{lakshminarayanan2017simple} are state-of-the-art uncertainty estimation methods that have demonstrated high-quality uncertainty estimation. Both MC-dropout (Fig. \ref{fig:arch_detail}a) and deep ensembles (Fig. \ref{fig:arch_detail}b) rely on multiple forward passes through their network(s), making inference inefficient when directly applied to SNNs \cite{sun2023efficient}. This inefficiency stems from the need for SNNs to propagate through all their layers multiple times to achieve accurate predictions for each input. Moreover, uncertainty estimation for regression tasks faces additional challenges due to the absence of existing solutions for SNN-based regression and the inherent difficulty of event-based SNNs to create continuous outputs, further complicating the process.

Previous research has examined uncertainty estimation for SNNs using MC-dropout in classification tasks \cite{sun2023efficient}. As illustrated in Fig \ref{fig:arch_detail}c,  the Average-Over-Time SNN (AOT-SNN) framework \cite{sun2023efficient} exploits the temporal processing capabilities of SNNs to reduce the computational cost of multiple forward passes, while still maintaining precise uncertainty estimation. Yet, unlike DNNs, SNNs inherently generate event-based discrete outputs rather than continuous ones, which adds a layer of complexity to adapt the AOT-SNN framework and other DNN-based uncertainty estimation techniques to regression tasks.


For classification tasks, DNNs typically compute a value for each class, which is subsequently converted into a normalized probability using the softmax function. In regression, estimating uncertainty is more challenging because a regular regression model only predicts a point estimate of the target variable as the outcome, rather than a probability distribution conditioned on an input. Deep ensembles \cite{lakshminarayanan2017simple}, often regarded as 'the gold standard for accurate and well-calibrated predictive distributions' \cite{wilson2020bayesian}, address this challenge by estimating both the mean and variance for a given input and subsequently generating a conditional probability distribution (CPD) of the target variable, assuming a heteroscedastic Gaussian distribution \cite{nix1994estimating}. Mixture density networks \cite{Bishop1994} predict more flexible CPDs through the mixtures of Gaussians assumption, with the mixing coefficients and parameters of the individual Gaussians being output by a neural network. 

Recently, the regression-as-classification (RAC) approach, which reformulates regression as a classification problem, has demonstrated excellent performance in regression tasks \cite{stewart2023regression}. Specifically, RAC divides the range of the target variable into equal-interval bins, treating each bin as a class \cite{guha2024conformal}. The RAC approach can often outperforms regular regression models \cite{stewart2023regression} and has been applied to various regression tasks, such as image colorization \cite{zhang2016colorful},
depth estimation \cite{fu2018deep}, age estimation \cite{rothe2015dex}, and pose estimation \cite{rogez2017lcr}. However, the discussion of DNN-based uncertainty estimation of the RAC approach is very limited: an RAC-based method is used to obtain conformal prediction sets for regression in \cite{guha2024conformal}, and in \cite{yu2022monocular}, deep ensembles or an auxiliary-network based method is proposed to estimate the uncertainty of depth estimation. To the best of our knowledge, there have been no discussions in the literature on uncertainty estimation for SNNs in regression using the RAC approach.


This paper presents two approaches for adapting the AOT-SNN framework to regression tasks. The first method, shown in Fig. \ref{fig:arch_detail}d, involves SNNs that generate both the mean and variance at each time step. In this approach, the predicted mean and variance are determined by all these time step outputs, similar to how AOT-SNN functions in classification tasks. The CPD of the target variable is computed by treating the output as a sample drawn from a (heteroscedastic) Gaussian distribution, characterized by the predicted mean and variance. The second method is based on the RAC approach, which allows the application of classification-based AOT-SNNs for uncertainty estimation. Under the assumption of a uniform distribution within each bin, this method uses the probabilities of the bins to predict CPDs. Both methods were validated using a toy dataset as well as experiments on standard benchmark regression datasets. The results demonstrate that both proposed SNN methods achieve uncertainty performance on par with state-of-the-art DNN approaches.


\section{Background}
In this section, we first formalize the regression problem from a probabilistic perspective. Then, we briefly introduce two approaches for probabilistic regression: the heteroscedastic Gaussian approach \cite{nix1994estimating,lakshminarayanan2017simple}, and the RAC approach \cite{stewart2023regression}.

\subsection{Problem Setup}
A regression task aims to predict a continuous target variable $y$ given an input vector $\mathbf{x}$. Formally, let $D = \{ (\mathbf{x}_i, y_i) \}_{i=1}^N$ denote a dataset of $N$ independent observations, where $\mathbf{x}_i \in \mathbb{R}^d$ represents the $d$-dimensional vector for the $i$-th observation, and $y_i \in \mathbb{R}$ is the corresponding target value. A regression model learns a mapping function $f: \mathbb{R}^d \rightarrow \mathbb{R}$ that minimizes a loss function, typically the mean squared error (MSE):
\begin{equation} 
\text{MSE} = \frac{1}{N} \sum_{i=1}^N (y_i - f(\mathbf{x}_i))^2.
\end{equation}

If we assume that given an input $\mathbf{x}$ in the training dataset $D$, the corresponding target value follows a (homoscedastic) Gaussian distribution that has a constant variance, the MSE loss is equivalent to a negative log-likelihood (NLL) loss.

\begin{figure*}[!t]
  \centering
     \includegraphics[width=0.85\textwidth]{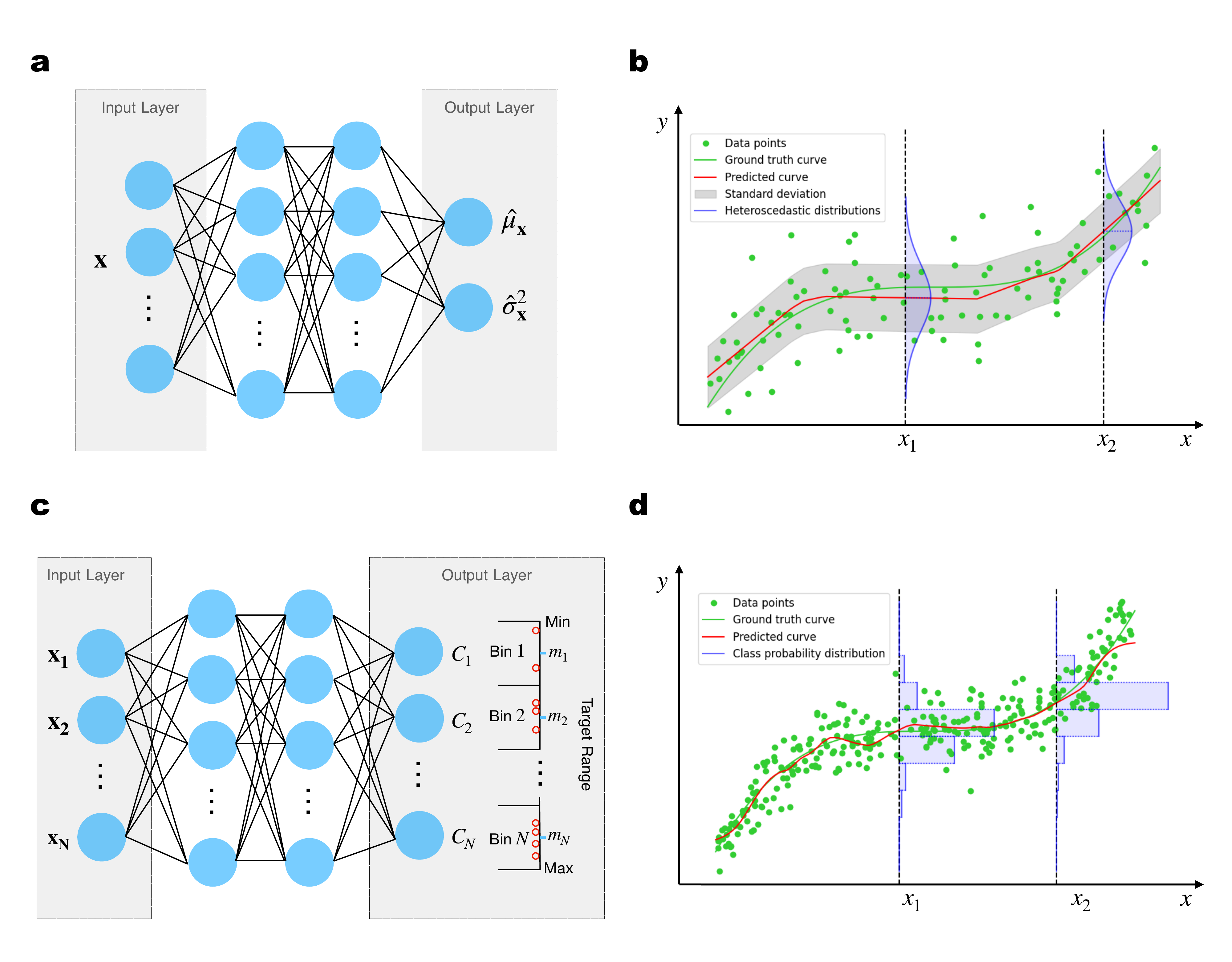}
  \caption{Illustration of the heteroscedastic Gaussian and Regression-as-Classification (RAC) approaches for uncertainty estimation in regression tasks. (a) Neural network architecture for predicting mean and variance. (b) Data fitting with the heteroscedastic Gaussian assumption. (c) RAC approach, discretizing the target variable into bins and predicting conditional probabilities. (d) Data fitting with the RAC approach under the uniform distribution assumption.}

  \label{fig:approaches}
\end{figure*}

\subsection{Regression with the heteroscedastic Gaussian assumption}
The constant variance assumption in the homoscedastic Gaussian distribution of the MSE loss may not be satisfied in many real regression problems. To tackle this issue, as depicted in Fig. \ref{fig:approaches}a, a neural network is commonly employed to compute two values: the mean $\mu(\mathbf{x})$ and the variance $\sigma^2(\mathbf{x})$ of the target variable $y$ \cite{nix1994estimating}. In this framework, $\mu(\mathbf{x})$ and $\sigma^2(\mathbf{x})$ are assumed to follow heteroscedastic Gaussian distributions (Fig. \ref{fig:approaches}b). The CPD of the corresponding target value $y$ can be written as:
\begin{equation} \mathbf{p}(y | \mathbf{x}) = \mathcal{N}(\mu(\mathbf{x}), \sigma^2(\mathbf{x})) \end{equation}.
The NLL loss used to train such a regression model is
\begin{equation} 
\text{NLL} = \frac{1}{2} \log(2\pi\sigma^2(\mathbf{x})) + \frac{(y - \mu(\mathbf{x}))^2}{2\sigma^2(\mathbf{x})}.
\end{equation}

Deep ensembles \cite{lakshminarayanan2017simple} carry out uncertainty estimation in regression based on this method. Specifically, deep ensembles train a specific number ($M$) of DNNs with identical architecture (Fig. \ref{fig:arch_detail}b). For an input $\mathbf{x}$, each of those DNNs outputs a mean $\mu_m(\mathbf{x})$ and a variance $\sigma_m^2(\mathbf{x})$. The predictive mean $\mu_*(\mathbf{x})$ and variance $\sigma_*(\mathbf{x})^2$ of an ensemble are written as:
\begin{align}
y_{*}(\mathbf{x}) &= \mu_*(\mathbf{x}) = \frac{1}{M}\Sigma_m\mu_m(\mathbf{x}). \\
\sigma_*^2(\mathbf{x}) &= \frac{1}{M}\Sigma_m(\sigma^2_m(\mathbf{x})+\mu^2_m(\mathbf{x})) - \mu^2_*(\mathbf{x}).
\end{align}

\subsection{Regression-as-Classification (RAC)}

As shown in Fig. \ref{fig:approaches}c, the RAC approach \cite{stewart2023regression} discretizes the range space $[y_{min}, y_{max}]$ of the target variable in a regression task into $K$ equal-sized intervals, called \textit{bins}. The boundaries and midpoints of bins are written as $b_1, b_2, \cdots, b_{K+1}$ and $m_1, m_2, \cdots, m_K$, respectively, with 
\begin{equation}
m_k = \frac{b_k + b_{k+1}}{2}. \label{equ:mi}
\end{equation}

Treating each bin as a class $C_k$, where $k=1, 2, \cdots, K$, a regression problem can be reformulated as a classification problem. The continuous target variables can be converted into discrete class labels $\Tilde{y}_i=\text{argmin}_j|y_i-m_j|$, where the smallest $j$ is taken in case of ties. By minimizing the cross-entropy loss, a neural network can be trained for the reformulated classification problem with the discretized dataset $\Tilde{D} = \{ (\mathbf{x}_i, \Tilde{y}_i) \}_{i=1}^N$. The RAC approach facilitates uncertainty estimation, as RAC models yield a normalized CPD for each input, written as $p_1, p_2, \cdots, p_K$, via the Softmax function. The predicted target value can be computed as the expected value over those probabilities $y_{*}(\mathbf{x}) = \sum_k p_km_k $.  

 \subsection{Quality of Uncertainty Estimation}
Calibration measures statistical compatibility of predictive probability distributions and real frequencies. Usually, calibration is taken as the indication of uncertainty quality \cite{lakshminarayanan2017simple}. A class of metrics to measure calibration is referred to as \textit{proper scoring rules} \cite{gneiting2007strictly}, which include the Brier score (BS) and negative log-likelihood (NLL). When evaluating the quality of probabilities, an optimal score output by a proper scoring rule indicates a perfect prediction, which means that it is not generated by trivial solutions. Note that proper scoring rules are usually taken as the loss function to train models. We evaluate model performance in terms of the Root Mean Squared Error (RMSE) and NLL. RMSE measures the accuracy of the predicted values, while NLL assesses the quality of the uncertainty estimates. 

\section{Method}
In this section, we introduce the AOT-SNN method and explain how we apply it to the heteroscedastic Gaussian regression model and the RAC model.

\subsection{AOT-SNN Method and Quality of Uncertainty Estimation}
SNNs typically use similar types of network topologies as DNNs, but their computation is distinct. SNNs employ stateful and binary-valued spiking neurons, as opposed to the stateless, analog-valued neurons of DNNs. Consequently, unlike the synchronous computation in DNNs, inference takes the form of an iterative process through multiple time steps $t = 0, 1, \dots, T$ in typical SNN models. During each time step $t$, the membrane potential of a spiking neuron $U_t$ is influenced by the impinging spikes from connected neurons emitted at time step $t-1$ and the previous membrane potential $U_{t-1}$. When the membrane potential $U_t$ reaches a threshold $\theta$, the neuron emits a spike. This sparse, asynchronous communication between connected neurons is key to enabling SNNs to achieve high energy efficiency.

MC-dropout \cite{yarin2016uncertainty} is a popular technique for estimating predictive uncertainty in DNNs. It involves enabling dropout during both training and inference, where for a given input, multiple forward passes are used to obtain a distribution of outputs. Applying this method directly to an SNN can be computationally inefficient due to its inherent time-step mechanism, which requires running the SNN multiple times for a single forward pass. As a result, multiple forward passes substantially increase resource demands for uncertainty estimation. In \cite{sun2023efficient}, an efficient framework named \textit{AOT-SNN} is introduced that leverages the time-step mechanism of SNNs for MC-dropout. Instead of running multiple forward passes, AOT-SNNs compute predictive distributions by averaging the outputs across time steps within a single forward pass, as illustrated in Fig. \ref{fig:arch_detail}c. This approach significantly reduces computational costs while still enhancing the quality of uncertainty estimation. Additionally, AOT-SNNs introduce the average-over-time loss function, which calculates loss by averaging over multiple time steps rather than using only the final time step:
\begin{equation}
L = \frac{1}{T}\sum_{t=1}^T l(t).
\end{equation}

\subsection{AOT-SNN over the heteroscedastic Gaussian model}
As illustrated in the Fig. \ref{fig:arch_detail}d, each time step of an AOT-SNN regression model with the heteroscedastic Gaussian assumption has two outputs, representing the the mean $\mu_t(\mathbf{x})$ and the variance $\sigma_t^2(\mathbf{x})$ of the target variable, respectively. In implementation, a readout integrator layer that has two non-spiking neurons is used as the output layer, as in \cite{yin2021accurate}. As such, the membrane potentials of these two neurons are taken as $\mu_t(\mathbf{x})$ and $\sigma^2_t(\mathbf{x})$. The predicted mean and variance for an sample $\mathbf{x}$ are given by $\mu_*(\mathbf{x}) = \frac{1}{T}\Sigma_t\mu_t(\mathbf{x})$ and $\sigma^2_*(\mathbf{x}) = \frac{1}{T}\Sigma_t(\sigma^2_t(\mathbf{x})+\mu^2_t(\mathbf{x})) - \mu^2_*(\mathbf{x})$.



\subsection{AOT-SNN over RAC}
\subsubsection{Predictive value of the target variable} 
As discussed previously, the RAC approach assigns a probability $p_i$ to each discretized bin, where $i = 1, 2, \cdots, K$. For uncertainty estimation, however, we need a probability distribution $\mathbf{p}(y | \mathbf{x})$ that spans the entire range of the continuous target variable $y$. To achieve this, we assume the predicted target value follows a uniform distribution within each bin (Fig. \ref{fig:approaches}d), represented as $f_k$, which can be formulated as
\begin{equation}
f_k = \frac{p_k}{b_{k+1}-b_k}. \label{equ:fi}
\end{equation}

With this setup, the expected value $\mathbb{E}[y]$ of the target variable can be calculated as follows:
\begin{align}
\mathbb{E}[y] &= \Sigma_i \int_{b_k}^{b_{k+1}} f_i y \, dy \\
              &= \Sigma_i f_i  \int_{b_k}^{b_{k+1}} y \, dy\\
              &= \Sigma_i f_i \cdot \left( \frac{b_{k+1}^2}{2} - \frac{b_k^2}{2} \right) \\
              &= \Sigma_i f_i \cdot \left( b_{k+1} - b_k \right) \left( \frac{b_{k} + b_{k+1}}{2} \right).
\end{align}

By substituting Equations \ref{equ:mi} and \ref{equ:fi}, we obtain:
\begin{equation}
\mathbb{E}[y] = \Sigma_k p_k \cdot m_k.
\end{equation} 
We take $\mathbb{E}[y]$ as the predictive value $y_{*}(\mathbf{x})$ for regression.

\subsubsection{Distance loss} 
To estimate uncertainty effectively, it is crucial to assign accurate probabilities not only to the ground-truth bin but also to its neighboring bins. Specifically, the bins closer to the ground-truth should be assigned higher probabilities. A model trained using cross-entropy loss does not necessarily achieve this. Therefore, we employ the loss function introduced in \cite{guha2024conformal}, referred to here as the \textit{distance loss}. The distance loss not only encourages the model to concentrate its predictions on bins closer to the actual values, thereby preserving the ordering of the output space, but also ensures that the model does not become overly confident in its predictions, leading to more reliable uncertainty estimation. In implementation, this loss function consists of two components:

\paragraph{Penalization for Distance} This component imposes a penalty on the predicted density for output bins that are distant from the true output value. It is defined as
\begin{equation}
\mathcal{L}_\text{dis} = \sum_{k=1}^{K} |k-j|^qp_k,
\end{equation}
where $q>0$ is a hyperparameter.

\paragraph{Entropy Regularization} This component encourages variability in the predictions by regularizing the model's output distribution. It is written as
\begin{equation}
\mathcal{H} = - \sum_{k=1}^{K} p_k\text{log}p_k.
\end{equation}

The total loss is defined as
\begin{equation}
\mathcal{L} = \mathcal{L}_\textbf{dis} + \tau \mathcal{H},
\end{equation}
where $\tau>0$ is a hyperparameter.

\subsection{Network configuration}
Similar to AOT-SNN over the heteroscedastic Gaussian model, the output layer consists of a readout integrator layer with $K$ non-spiking neurons, corresponding to $K$ classes. At each time step, the membrane potentials of these non-spiking neurons are treated as logits, which are then converted to probabilities using the Softmax function. The final prediction is obtained by averaging the probabilities across all time steps, as in AOT-SNNs for classification \cite{sun2023efficient}.

\section{Experiments}
In this section, we describe the experimental evaluation of the proposed AOT-SNN methods for regression. We validate our approaches on both a toy dataset and the standard benchmark regression datasets to assess their uncertainty estimation capabilities in comparison with DNN methods.

\subsection{Toy Dataset}

\begin{figure}[htb]
  \centering
     \includegraphics[width=0.5\textwidth]{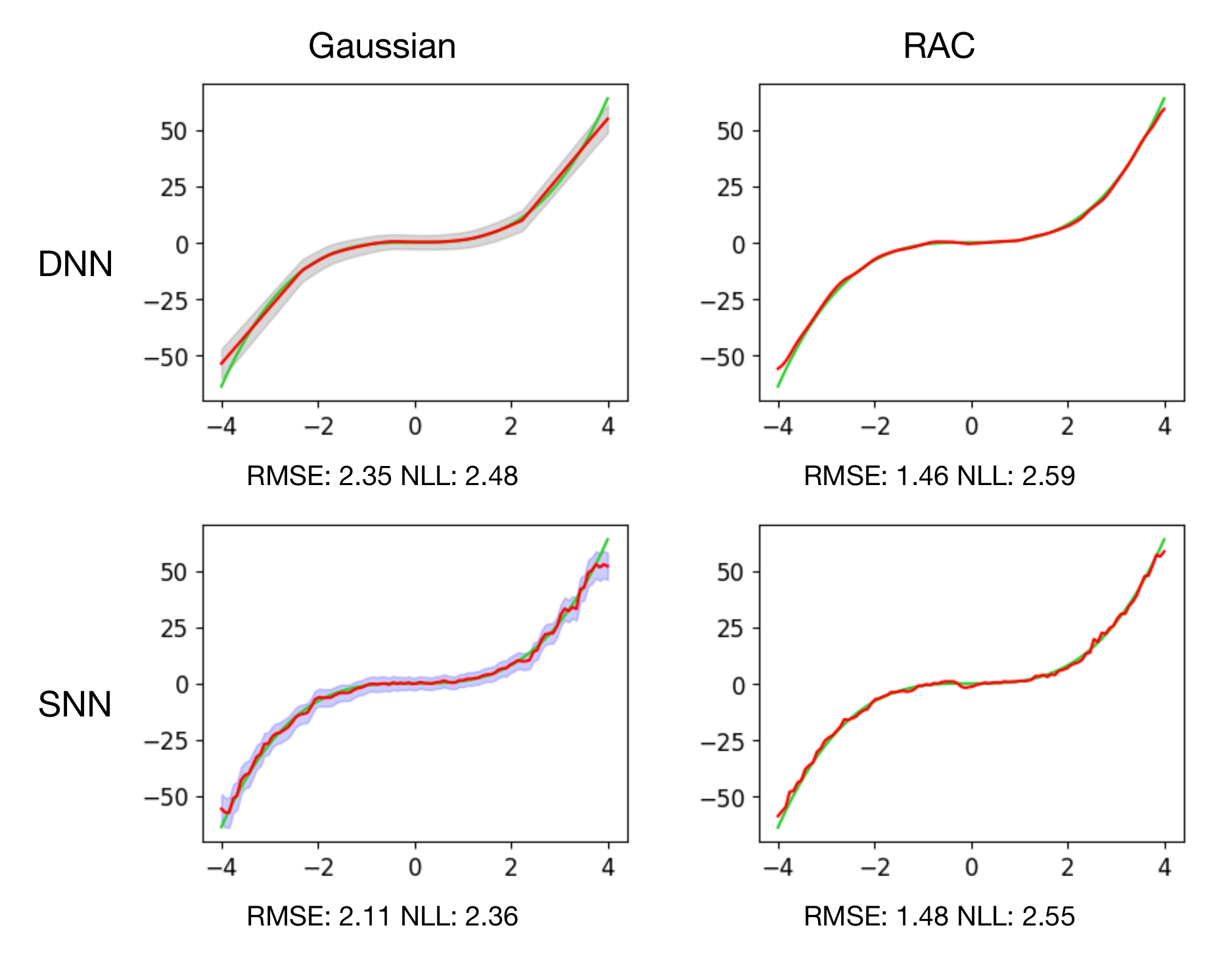}
  \caption{Comparison of RMSE and NLL metrics for AOT-SNN and DNN models using Gaussian and RAC approaches on the toy dataset. The Gaussian-based AOT-SNN model shows improved RMSE and NLL compared to its DNN counterpart, while the RAC-based AOT-SNN achieves comparable results to the RAC DNN model.}
  \label{fig:toy}
\end{figure}

\newcommand{\tpm}{$\pm$}
\begin{table*}[h]%
\begin{center}
\caption{Comparisons of RMSE on regression benchmark datasets. Dataset size ($N$) and input dimensionality ($Q$) are also listed. Better results between the Gaussian and RAC models are highlighted in bold.}
\label{tab:rmse}
\resizebox{2.\columnwidth}{!}{
\begin{tabular}{lcc | cc | cc}
\toprule 
 &&& \multicolumn{2}{|c|}{DNNs} & \multicolumn{2}{c}{AOT-SNNs} \\
Datasets & $N$ & $Q$ & MC-dropout \cite{yarin2016uncertainty} & Deep Ensembles \cite{lakshminarayanan2017simple} &   Gaussian & RAC \\ 
 \midrule
Boston housing & 506 & 13  &  \textit{2.97 $\pm$ 0.19}  & \textit{3.28 $\pm$ 1.00}  & 3.35 \tpm 0.17 &  3.40 \tpm 0.19 \\
Concrete strength & 1,030 & 8   & \textit{5.23 $\pm$ 0.12}  & \textit{6.03 $\pm$ 0.58}  & 5.79 \tpm 0.11 & \textbf{5.08 \tpm 0.10}  \\
Energy efficiency  & 768 & 8 &  \textit{1.66 $\pm$ 0.04}  & \textit{2.09 $\pm$ 0.29}  &  2.09 \tpm 0.13 & \textbf{1.26 \tpm 0.02} \\
Kin8nm    & 8,192 & 8 &  \textit{0.10 $\pm$ 0.00}  & \textit{0.09 $\pm$ 0.00}  &  0.08 \tpm 0.00 & 0.08 \tpm 0.00  \\
Naval propulsion  & 11,934 & 16 &  \textit{0.01 $\pm$ 0.00}  & \textit{0.00 $\pm$ 0.00}  &  0.01 \tpm 0.00 & \textbf{0.00 \tpm 0.00}  \\
Power plant   & 9,568 & 4 & \textit{4.02 $\pm$ 0.04} & \textit{4.11 $\pm$ 0.17}  & 5.13 \tpm 0.08 & \textbf{4.39 \tpm 0.03}  \\
Protein structure  & 45,730 & 9 &  \textit{4.36 $\pm$ 0.01}  & \textit{4.71 $\pm$ 0.06}  &  4.31 \tpm 0.02 & \textbf{4.13 \tpm 0.03}  \\
Wine quality red   & 1,599 & 11 & \textit{0.62 $\pm$ 0.01}  & \textit{0.64 $\pm$ 0.04}  & \textbf{0.62 \tpm 0.01} & 0.64 \tpm 0.01 \\
 \bottomrule
\end{tabular}
}
\end{center}
\end{table*}

\begin{table*}[h]%
\caption{Comparisons of NLL on regression benchmark datasets. Dataset size ($N$) and input dimensionality ($Q$) are also listed. Better results between the Gaussian and RAC models are highlighted in bold.}
\label{tab:nll}
\begin{center}
\resizebox{2.0\columnwidth}{!}{
\begin{tabular}{lcc | cc | cc}
\toprule 
 &&& \multicolumn{2}{|c|}{DNNs} & \multicolumn{2}{c}{AOT-SNNs} \\
Datasets & $N$ & $Q$ &  MC-dropout \cite{yarin2016uncertainty} & Deep Ensembles \cite{lakshminarayanan2017simple} & Gaussian & RAC \\ 
 \midrule
Boston housing & 506 & 13  & \textit{2.46 $\pm$ 0.06}  & \textit{2.41 $\pm$ 0.25} & 2.74 \tpm 0.04 &  \textbf{2.59 \tpm 0.02}   \\
Concrete strength   & 1,030 & 8 &  \textit{3.04 $\pm$ 0.02}  & \textit{3.06 $\pm$ 0.18} & 3.17 \tpm 0.02 & \textbf{3.08 \tpm 0.01} \\
Energy efficiency  & 768 & 8 &  \textit{1.99 $\pm$ 0.02}  & \textit{1.38 $\pm$ 0.22} & 1.93 \tpm 0.08 & \textbf{1.52 \tpm 0.01} \\
Kin8nm   & 8,192 & 8 &  \textit{-0.95 $\pm$ 0.01}  & \textit{-1.20 $\pm$ 0.02} & \textbf{-1.13 \tpm 0.00} & -0.75 \tpm 0.00 \\
Naval propulsion  & 11,934 & 16   & \textit{-3.80 $\pm$ 0.01}  & \textit{-5.63 $\pm$ 0.05} &  -3.76 \tpm 0.23 & \textbf{-4.30 \tpm 0.00}\\
Power plant   & 9,568 & 4 & \textit{2.80 $\pm$ 0.01}  &\textit{2.79 $\pm$ 0.04} & 3.19 \tpm 0.02 & \textbf{3.05 \tpm 0.00}\\
Protein structure & 45,730 & 9 & \textit{2.89 $\pm$ 0.00}  & \textit{2.83 $\pm$ 0.02} & 2.79 \tpm 0.03 & \textbf{2.32 \tpm 0.01} \\
Wine quality red  & 1,599 & 11 & \textit{0.93 $\pm$ 0.01}  & \textit{0.94 $\pm$ 0.12} &  1.28 \tpm 0.04 & \textbf{0.21 \tpm 0.03} \\
 \bottomrule
\end{tabular}
}
\end{center}
\end{table*}

The toy dataset consists of 100 training examples randomly drawn from $y = x^3 + \epsilon$, where $x \in [-4, 4]$ and $\epsilon \sim \mathcal{N}(0, 3^2)$. We have 100 evenly spaced numbers over $[-4, 4]$ for the testing examples. Both the regression with a heteroscedastic Gaussian assumption (referred to as the \textit{Gaussian} approach) and the RAC approach were applied to the MC-dropout-based AOT-SNNs. Regular DNNs, used as a baseline, and the SNNs each had a single hidden layer of 100 neurons. The DNNs employed the ReLU activation function. While Gaussian models were trained using the NLL loss, RAC models were trained using the distance loss with $\tau=1$, optimized via the grid search. Both SNN models set their dropout rate as 0.05. The neuron model used was PLIF \cite{fang2021incorporating}, where the time constant is learned and shared within the same layer. The number of time steps for the SNN models was set to 8. For AOT-SNN over RAC, we constructed an SNN with $K=150$ outputs, corresponding to 150 classes, also optimized through grid search. 

Fig. \ref{fig:toy} presents the results on the toy dataset. The RAC approach achieves better RMSEs, which is consistent with the observations in \cite{stewart2023regression}. In contrast, the Gaussian approach slightly outperforms the RAC approach in NLL, likely due to the use of the NLL loss function. Comparing our AOT-SNN models to the baseline DNN models, the Gaussian-based AOT-SNN significantly outperforms its DNN counterpart in both RMSE and NLL, while the RAC-based AOT-SNN achieves comparable performance to the RAC DNN model. The strong performance of our AOT-SNN models can be attributed to the AOT-SNN framework, where MC-dropout is applied throughout the time steps of SNN models.

\subsection{Benchmark Datasets}
To compare the predictive performance of our AOT-SNN models with other state-of-the-art methods, we conducted experiments on the standard benchmark datasets from the UCI Machine Learning Repository. The dataset size ($N$) and input dimensionality ($Q$) are provided in Tables \ref{tab:rmse} and \ref{tab:nll}. Each dataset was split into 20 train-test folds, except for the protein dataset, which was divided into 5 folds. For each fold, we selected the dropout rate from the set $[0.005, 0.01, 0.05, 0.1]$ that performed best in terms of NLL on the validation set, consistent with the approach in \cite{yarin2016uncertainty}. In AOT-SNNs, we used a network with a hidden layer of 200 PLIF neurons. The Gaussian models were trained for 600 epochs with the NLL loss, while the RAC models were trained for 200 epochs using the distance loss with $\tau=1$. For the RAC models, we set $K=50$ for the output layer, following the choice made in \cite{guha2024conformal} for the 	\textit{Concrete Strength} and 	\textit{Energy Efficiency} datasets. We found that this value provided satisfactory performance on these benchmark datasets. Further refinement of this hyperparameter could be explored to potentially enhance model performance for each dataset.

The RMSE and NLL results for the benchmark datasets are summarized in Tables \ref{tab:rmse} and \ref{tab:nll}, respectively. The RAC-based AOT-SNN models consistently demonstrate strong performance in terms of RMSE, with significant improvements observed for the \textit{Concrete Strength}, \textit{Energy Efficiency}, and \textit{Protein Structure} datasets. Meanwhile, the AOT-SNN Gaussian model performs comparably to MC-dropout and Deep Ensembles for most datasets in terms of RMSE. For NLL, both the AOT-SNN RAC model and the AOT-SNN Gaussian model perform comparably to MC-dropout and Deep Ensembles on most datasets. When comparing the two approaches within AOT-SNNs, the RAC models outperform the Gaussian models on most datasets, highlighting the effectiveness of the RAC approach for uncertainty estimation in regression tasks.




\section{Conclusion}
In this paper, we presented two approaches for adapting the AOT-SNN framework to regression tasks: one based on the assumption of heteroscedastic Gaussian distributions and the other leveraging the RAC approach. Our results demonstrate that both approaches achieve strong performance in terms of uncertainty estimation, often surpassing traditional DNN-based methods in RMSE and NLL metrics, particularly in the case of RAC-based models. By exploiting the temporal dynamics of spiking neurons and incorporating uncertainty estimation through the AOT-SNN framework, our proposed methods provide a novel, energy-efficient solutions for uncertainty estimation in regression tasks. Future work will focus on further optimizing these models for hardware implementations, thereby enabling practical deployment in real-time, resource-constrained environments.

\bibliographystyle{IEEEtran}
\bibliography{ref}

\end{document}